\documentclass[letterpaper, 10 pt, conference]{ieeeconf}  

\usepackage[pdftex,dvipsnames]{xcolor}  
\usepackage{pgfplots}
\usepackage{tikz}
\pgfplotsset{compat=newest}
\usepackage[colorinlistoftodos,prependcaption,textsize=tiny]{todonotes}

\IEEEoverridecommandlockouts                              
\renewcommand{\baselinestretch}{0.96}

\usepackage{amsthm}
\usepackage{graphics} 
\usepackage{epsfig} 
\usepackage{times} 
\usepackage{amsmath} 
\usepackage{amssymb}  

\usepackage[normalem]{ulem}

\usepackage[maxbibnames=99,giveninits=true]{biblatex}
\usepackage{bm}
\usepackage{bbm}
\usepackage{xcolor}
\usepackage{graphicx}
\usepackage{algorithm}
\usepackage[noend]{algpseudocode}
\usepackage{booktabs}
\usepackage{siunitx}
\usepackage{hyperref}

\usepackage{caption}
\usepackage{lipsum}
\usepackage{multirow}
\usepackage{diagbox} 
\usepackage{subcaption}

\newcommand{\R}{\mathbb{R}}

\newtheorem{definition}{\bf Definition}

\newtheorem{theorem}{\bf Theorem}
\newtheorem{lemma}{\bf Lemma}

\newtheorem{remark}{\bf Remark}

\begin{document}
\title{\LARGE \bf
Scalable Networked Feature Selection with Randomized Algorithm \\ for Robot Navigation }
\author{Vivek Pandey*, Arash Amini*, Guangyi Liu, Ufuk Topcu, Qiyu Sun, Kostas Daniilidis, and Nader Motee
\thanks{
 $*$ Vivek Pandey and Arash Amini contributed equally to this work. \endgraf V. Pandey, G. Liu and N. Motee are with the Department of Mechanical Engineering and Mechanics, Lehigh University. {\tt\small \{vkp219,gliu,motee\}@lehigh.edu}.\endgraf A. Amini and U. Topcu are with the Department of Aerospace Engineering and Engineering Mechanics, The University of Texas at Austin. {\tt\small \{a.amini,utopcu\}@utexas.edu}.\endgraf
 Q. Sun is with the Department of Mathematics, University of Central Florida. {\tt\small qiyu.sun@ucf.edu}.\endgraf
 K. Daniilidis is with the Department of Computer and Information Sciences, University of Pennsylvania. {\tt\small kostas@cis.upenn.edu}.\endgraf
}
}

\maketitle

\thispagestyle{plain}
\pagestyle{plain}

\begin{abstract} 

We address the problem of sparse selection of visual features for localizing a team of robots navigating an unknown environment, where robots can exchange relative position measurements with neighbors. We select a set of the most informative features by anticipating their importance in robots localization by simulating trajectories of robots over a prediction horizon. Through theoretical proofs, we establish a crucial connection between graph Laplacian and the importance of features. We show that strong network connectivity translates to uniformity in feature importance, which enables uniform random sampling of features and reduces the overall computational complexity. We leverage a scalable randomized algorithm for sparse sums of positive semidefinite matrices to efficiently select the set of the most informative features and significantly improve the probabilistic performance bounds.  
Finally, we support our findings with extensive simulations.

\end{abstract}


\section{Introduction}\label{sec:Intro}
Navigation is a critical component of robotics, enabling robots to perform various tasks autonomously. To achieve successful navigation, robots rely on continuous position estimation or localization relative to a specific frame of reference. To obtain a good estimate of their position, robots rely on various sensors like LiDAR \cite{guadagnino2022fast_sparse_lidar}, and algorithms like Kalman filtering \cite{thrun2005_probabilistic_robotics}. 

A key challenge in robot navigation is achieving accurate localization with provable performance bounds while minimizing computational costs due to onboard power limitations. Extensive research has addressed this challenge, focusing on efficient methods for robot localization \cite{sala2006landmark_selection_pose,strasdat2009landmark_useful,lerner2007landmark_selection_task,gorbenko2012landmark_selection_minimalset}. These methods typically involve selecting informative features based on the robot's current position and projected trajectory. The robot's position is then estimated by fusing information from these selected landmarks. In \cite{thrun2004info_filter,thrun2003info_filter_multi_robot_slam}, the authors utilize computationally efficient sparse extended information filters for simultaneous localization and mapping (SLAM), avoiding the computationally expensive matrix inversion step. In \cite{carlone2019attention}, the authors leverage a linear motion and vision model to track features along the robot's predicted trajectory. They then select the most informative features using the greedy method and provide performance bounds using the submodularity results in \cite{nemhauser1978maximizing_submodular_Set_functions}. The authors in \cite{zhao2020good_feature_matching} propose a solution to the latency problem in Visual SLAM by combining efficient feature selection with active feature matching. In \cite{jiao2022lidar_slam_feature_selection}, the authors utilize the stochastic greedy algorithm proposed in \cite{mirzasoleiman2015stochastic_greedy} for feature selection for LiDAR SLAM. In \cite{hossein2020feature_fast}, the authors propose a randomized algorithm that offers probabilistic performance guarantees while improving the algorithmic complexity of the feature selection problem. Feature selection methods traditionally targeted single agents, but networked systems like consensus  networks \cite{Hossein2020characterization_performance} demand techniques suited for interconnected systems.

Building upon the sparse feature selection \cite{hossein2020feature_fast}, we address the localization problem for a team of robots navigating an unknown environment. These robots can observe  relative position  measurements with their neighbors and are assumed to follow a predefined trajectory. Each robot utilizes a camera to track selected features, aiming to enhance collective position estimation. 
We propose a randomized algorithm for feature selection in a multi-agent scenario where robots exchange relative measurement over a communication graph. To quantify the importance of features for localization, we simulate the trajectories of robots over a fixed prediction horizon and obtain information matrices corresponding to each feature. The information from the selected features is fused with the dynamical model to localize the team of robots at each time step over the prediction horizon. We employ recent findings on tail bounds for sums of random matrices from \cite{tropp2012tailbounds_random_matrix} to significantly enhance the probabilistic guarantees of our proposed algorithm. The randomized algorithm assigns probabilities to each feature based on the leverage scores of their information matrices, effectively measuring the information content of each feature. The algorithm samples features based on the probability distribution. We further analyze the crucial role of network connectivity in both feature selection problem and performance guarantees. 

{\it Our contribution:} We address the problem of sparse feature selection for localizing a team of agents, where they exchange relative measurements leading to a graphical network. Compared to \cite{hossein2020feature_fast}, we significantly improve the probabilistic bound of the randomized feature selection algorithm. By fusing the inter-agent measurement information, we establish crucial connection between graph Laplacian and leverage score of features. We show that strongly connected graphs leads to uniform feature importance, enabling efficient random sampling and reduced computational complexity. The proofs of all theoretical results are provided in Appendix


\section{Mathematical Notation}\label{sec:math_notation}

We denote a vector and matrix valued variable by lowercase and uppercase letters respectively. All vectors formed by stacking a sequence of vectors indexed over time is denoted by boldface lower case letters. Similarly, matrices formed by augmenting a sequence of matrices over time is represented by boldface uppercase.  
We reserve the notation $S^n_{+} (S^n_{++})$ to denote the cone of symmetric positive semidefinite (positive definite) $n \times n$ matrices . The Kronecker product of two matrices $X_1, X_2$ is denoted by $X_1\otimes X_2.$ 
For a matrix $X$, we denote its transpose by $X^T$. 
We denote the special orthogonal group in $\R^3$ by SO(3). The cardinally of a set $A$ is given by $|A|$. We use $\textup{Tr}(\cdot)$ to denote the trace of a matrix. We reserve the symbol $I_{(\cdot)}$ for Identity matrix whose size is usually clear from the context.  
For every $X_1, X_2 \in S^n_{++} $, we write $X_2 \preceq X_1$ if and only if $ X_1 - X_2 \in S^n_{+}$. A Gaussian random vector $z$ with mean $\mu$ and covariance matrix $\Sigma$ is denoted by $z \sim \mathcal{N}(\mu, \Sigma).$ The set of nonnegative integers and reals are denoted by $\mathbb{Z}_+$ and $\mathbb{R}_+$ respectively.

\vspace{0.1cm}
\noindent{\it  Graph Theory:} A weighted graph is defined by $\mathcal{G} = (\mathcal{V}, \mathcal{E}, \omega)$, where $\mathcal{V}$ is the set of nodes, $\mathcal{E}$ is the set of edges, and $\omega: \mathcal{V} \times \mathcal{V} \rightarrow \mathbb{R}_{+}$ is the weight function that assigns a non-negative number to every link. Two nodes are directly connected if and only if $(i,j) \in \mathcal{E}$.

The Incidence matrix of a directed graph $\mathcal{G}$ is a $|\mathcal{V}| \times |\mathcal{E}|$ matrix $C$ such that 
\[C_{ij} = \begin{cases}
-1 ~ \text{if edge $e_j$ leaves vertex $v_i$} \\
1 ~ \text{if edge $e_j$ enters vertex $v_i$} \\
0 ~ \text{otherwise} 
\end{cases}\]
The Laplacian matrix of the graph $\mathcal{G}$ is $L = CWC^T$, where $W$ is a $|\mathcal{E}| \times |\mathcal{E}|$ diagonal matrix containing the edge weights \cite{siami2018centrality_measures}.
\begin{definition}\label{def:monotone_map}
    A map $\rho: S^n_{++} \rightarrow \R$ is called monotonically decreasing if for every $X_2 \preceq X_1 \implies \rho(X_2) \geq \rho(X_1).$
\end{definition}

\section{Problem Statement}\label{sec:problems_tatement}

We consider a multi-agent navigation scenario where agents localize themselves using camera and inter-agent measurement communication. Due to the large number of visual features extracted from the camera, selecting a subset of informative features becomes necessary to address computational complexity. We address the problem of sparse feature selection to estimate the positions of a team of $N$ robots navigating in an unknown environment, where robots exchange inter-agent measurements with their neighbors in the communication graph. Using a randomized algorithm, we select a set of the most informative features by anticipating their importance over a fixed time horizon. We further analyze how the graph Laplacian affects performance measures and the relative importance of features.

Let $\bm{x}_{t} \in \R^{3N}$ be the state vector of the team of robots at time $t \in \mathbb{Z}_+$. Consider the vector of discrete time horizon given by $[t:t+M] = [t, t+1, \cdots, t+M]$. Then vector
$\textbf{x}_{t: t+M}$
which contains states of the team of robots for all $t \in [t:t+M]$ can be written as 
\begin{equation}\label{eqn: robot_state_vectors_stacked}
    \textbf{x}_{t: t+M} = \left[x_t^T, x_{t+1}^T, \cdots, x_{t+M}^T\right]^T \in \R^{3N\left(M+1\right)}.  
\end{equation}
Since the state vector is often corrupted by Gaussian noise, obtaining an accurate robot state estimate often requires state estimation techniques like the Kalman Filter. Kalman Filter is an optimal recursive algorithm, designed for linear dynamic systems for estimating the true state based on a series of noisy measurements.  The Kalman Filter's versatility makes it widely used in fields like control systems, navigation, robotics \cite{thrun2005_probabilistic_robotics}, where accurate state estimation from noisy data is critical.
The Kalman filter is a minimum-mean square estimator which minimizes the error between the true state and the predicted state. 

For the vector $\textbf{x}_{t: t + M}$
the algorithm yields mean vector $\Bar{\bm{\mu}}_{t:t + M} \in \R^{3N(M+1)}$, and covariance matrix
$\Bar{\bm{\Sigma}}_{t:t + M} \in S^{3N(M+1)}_{++}$, which are the measure of the best estimate of state and the estimated accuracy respectively. The Kalman Filter algorithm requires inverting the covariance matrix. In scenarios where the size of the covariance matrix is large, inverting the matrix can be computationally expensive. To avoid this, we utilize the framework of extended information filter proposed in \cite{thrun2004info_filter}, which updates information vector and information matrix instead of the mean vector and the covariance matrix. 

For the mean $\Bar{\bm{\mu}}_{t:t + M}$ and covariance matrix $\Bar{\bm{\Sigma}}_{t:t + M}$,
the corresponding information vector $\Bar{\textbf{b}}_{t: t + M}^T  \in \R^{3N(M+1)}$ and information matrix $\Bar{\textbf{H}}_{t: t + M} \in S^{3N(M+1)}_{++}$ for the robot's state $\textbf{x}_{t: t+M}$ can be written as 
\begin{equation}\label{eqn:info_vec__mat_def}
\begin{aligned}
    \Bar{\textbf{b}}_{t: t + M} &= \Bar{\bm{\mu}}_{t: t + M} \Bar{\bm{\Sigma}}_{t: t + M}^{-1},\\
    \Bar{\textbf{H}}_{t: t + M} &=  \Bar{\bm{\Sigma}}_{t: t + M}^{-1}.
    \end{aligned}
\end{equation}
The information filter and the Kalman filter exhibit a clear mathematical connection, as shown in \eqref{eqn:info_vec__mat_def}. This interchangeability makes information filters a valuable tool for feature selection, as previously explored in \cite{carlone2019attention, hossein2020feature_fast}. Information filters offer computational advantages because the information vector and matrix can be updated linearly by adding the contribution from new observations.

Let $\Theta_t$ represent the set of all features $f$ available at a given time step. Furthermore, let the vector 
$\textbf{b}_{t: t + M}^f$
and the matrix 
$\textbf{H}_{t: t + M}^f$
be the contribution of new features to the information matrix, then the information vector and information matrix are updated as follows
\begin{equation}\label{eqn:info_vec_mat_update_def}
\begin{aligned}
        \textbf{b}_{t: t + M}(\Theta_t) &= \Bar{\textbf{b}}_{t: t + M} + \sum_{f \in \Theta_t}^{}\textbf{b}_{t: t + M}^f,\\
    \textbf{H}_{t: t + M}(\Theta_t) &= \Bar{\textbf{H}}_{t: t + M} + \sum_{f \in \Theta_t}^{}\textbf{H}_{t: t + M}^f.
\end{aligned}
\end{equation}
Due to limitations in on-board computational resources, it is often desirable to select a smaller subset, denoted by $\Phi_t$, of features from the available set, $\Theta_t$. This subset should prioritize features that contribute significantly to the overall information matrix, maximizing the information gain while minimizing the computational burden.
  
Our primary aim in this work is to select informative features that collectively lead to enhanced position estimates for all robots within the team. To achieve this, we model the team as a network, fusing relative position measurements obtained by the robots with their individual visual observations within a comprehensive robot motion model. The subsequent section delves further into the details of each of these models.

\section{Model for Robot Motion, Relative Measurements and Vision System}\label{sec:models}

To achieve accurate position estimates for the robots, we continuously update the initial estimates obtained from the robots' dynamics using information gathered from sensor measurements. This section begins by introducing the robot motion model, which mathematically describes how the robots' positions evolve over time. This model lays the foundation for incorporating measurement models, which will be used to refine the position estimates based on sensor data.

\begin{figure}[t]
    \centering
    \includegraphics[width=\linewidth]{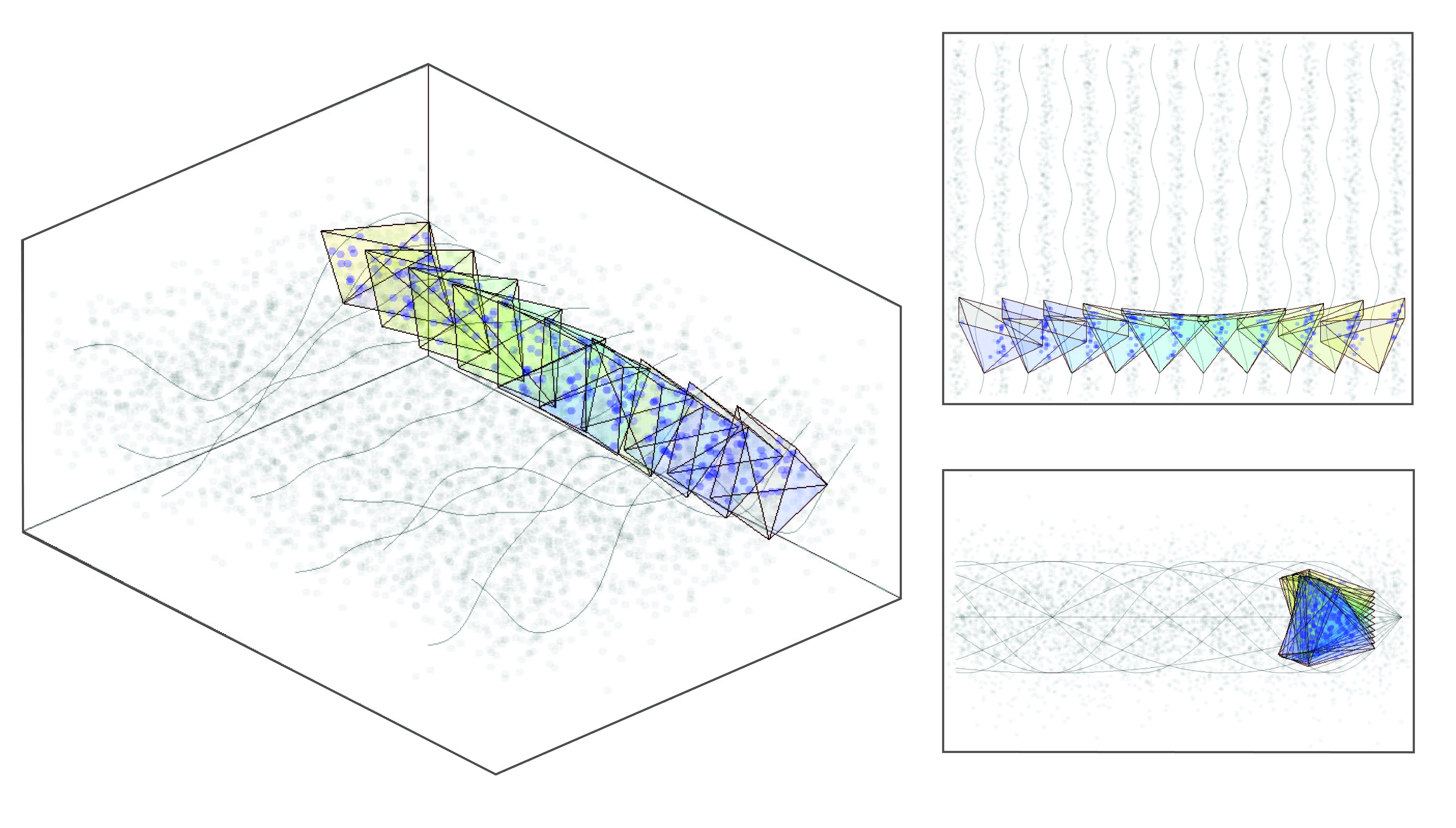}
    \caption{Description of robots and features in environment.}
    \label{fig:Robots_environment}
\end{figure}

\subsection{Model for Robots Motion}

We consider a team of $N$ robots such that the 
 dynamics of the $i$'th robot's motion is governed by the following linear dynamical model:
\begin{equation}\label{eqn:robot_dynamics_lin_ith}
    x_{\tau}^i = A x_{\tau-1}^i +  B u_{\tau}^i + \delta_{\tau}^i,
\end{equation}
for all $\tau \in [t:t+M]$, where $x_{\tau}^i$ is the state (position vector) of the $i$'th robot at time $\tau$, $ u_{\tau}^i$ is the control input of the $i$'th robot at time $\tau$ and $\delta_{\tau} ^i \sim \mathcal{N}(0, \Lambda_{\tau})$ such that $\mathbb{E}[\delta_{\tau_1}^i \delta_{\tau_2}^i] = 0$ for all $\tau_1 \neq \tau_2.$

Each robot moves along a pre-defined trajectory such that the control input for the $i$'th robot is given by  
\begin{equation}\label{eqn:control_input_ith_x}
    u_{\tau}^i = \bm{h}\left(u_{\tau}^{i,\text{ref}},x_{\tau -1}^i\right).
\end{equation}
Since the state of the $i$'th robot $x_{\tau -1}$ is a random variable, we apply the control law using its best estimate $\mu_{\tau -1}$ such that the control law becomes
\begin{equation}\label{eqn:control_input_ith_mu}
    u_{\tau}^i = \bm{h}\left(u_{\tau}^{i,\text{ref}},\mu_{\tau -1}^i\right).
\end{equation}
The overall dynamics of $N$ robots can be written by stacking the dynamics of each robot as 
\begin{equation} \label{eqn:robot_dynamics_team}
    \bm{x}_{\tau} = \bm{A} \bm{x}_{\tau -1} + \bm{B} \bm{u}_{\tau -1} + \bm{E}\bm{\delta}_{\tau},
\end{equation}
where $\bm{A} =\left(I_{N} \otimes A \right), \bm{B} =\left(I_{N} \otimes B \right)$ and $E = I_N \otimes I_3.$ 

Let $\Bar{\bm{\mu}}_t  = \mathbb{E}[\bm{x}_{\tau}]$ and $\Bar{\bm{\Sigma}}_t = \mathbb{E}[\left(\bm{x}_{\tau}-\Bar{\bm{\mu}}_t\right) \left(\bm{x}_{\tau}-\Bar{\bm{\mu}}_t\right) ^T]$, then the mean and covariance of $\textbf{x}_{t:t+M}$ can be expressed as 
\begin{equation}    \label{eqn:team_mu_mat_mu_tT}
    \Bar{\bm{\mu}}_{t: t + M} =     \begin{bmatrix}
        \Bar{\bm{\mu}}_{t}& \Bar{\bm{\mu}}_{t + 1} & \cdots & \Bar{\bm{\mu}}_{t + M}\\ 
    \end{bmatrix}
\end{equation}
\begin{equation}    \label{eqn:team_cov_mat_Sigma_tT}
\Bar{\bm{\Sigma}}_{t: t + M} = 
    \begin{bmatrix}
        \Bar{\bm{\Sigma}}_{t} & \Bar{\bm{\Sigma}}_{t,t+1} & \cdots & \Bar{\bm{\Sigma}}_{t,t+M} \\
        \Bar{\bm{\Sigma}}_{t,t+1}^T & \Bar{\bm{\Sigma}}_{t+1} & \cdots & \Bar{\bm{\Sigma}}_{t+1,t+M} \\
        \vdots & \vdots & \ddots & \vdots \\
        \Bar{\bm{\Sigma}}_{t,t+M}^T & \Bar{\bm{\Sigma}}_{t+1,t+M}^T & \cdots & \Bar{\bm{\Sigma}}_{t+M}  
    \end{bmatrix}
\end{equation}
where $\Bar{\bm{\Sigma}}_{\tau} = \bm{A}\Bar{\bm{\Sigma}}_{\tau - 1} \bm{A}^T + I_N \otimes \Lambda_{\tau}$ for all $\tau \in [t: t+M]$ and \[\Bar{\bm{\Sigma}}_{\tau_1, \tau_2} = \bm{A}^{(\tau_2 - \tau_1) }  \Bar{\bm{\Sigma}}_{\tau_1}\]
for all $\tau_1, \tau_2 \in [t: t+M]$ with $\tau_1 < \tau_2.$

The information matrix of $\textbf{x}_{t:t+M}$ using the dynamical model can be written as 
\begin{equation}\label{eqn:team_initial_info_mat}
    \Bar{\textbf{H}}_{t:t+M} = \Bar{\bm{\Sigma}}_{t: t + M}^{-1}.
\end{equation}
The information matrix $\Bar{\textbf{H}}_{t:t+M}$ is a measure of information content of $\textbf{x}_{t:t+M}$ when the system dynamics is predicted over the fixed time horizon $M.$ 
The following subsections delve into the process of sensor fusion. Here, we explore how information matrices, derived from the robot motion model, are dynamically updated by incorporating measurements from the robot's sensors.

\subsection{Information from Relative Measurements}

Although the robots have initial estimates of their positions based solely on predicted dynamics, we improve their estimates by fusing relative measurements of their positions. These measurements are obtained from neighboring agents within a connected network, where each robot can determine its relative position with respect to its neighbors in the communication graph $\mathcal{G}_{\tau}$. 

The relative measurement vector $ \xi_{\tau}^{i,j}$ at time $\tau$ between two such neighboring agents $i$ and $j$ can be expressed as 
\begin{equation}\label{eqn:relative_meas}
    \xi_{\tau}^{i,j} = x_{\tau}^i - x_{\tau}^j + \zeta_{\tau}^{i,j},
\end{equation}
where $\zeta^{i,j}_{\tau} \sim \mathcal{N}\left(\bm{0}, \text{P}_{\tau}\right).$

By stacking together the relative measurements for the entire network, the measurement equation can be written compactly as 
\begin{equation}\label{eqn:team_relative_meas}
    \bm{\xi}_{\tau} = (C_{\tau} \otimes I_{3})^T \bm{x}_{\tau}+ \bm{\zeta}_{\tau},
\end{equation}
where the matrix $C$ is the incidence matrix of the network and $\bm{\zeta}_{\tau} \sim \mathcal{N}(\bm{0}, \textbf{P}_{\tau})$. By choosing the weight matrix $W$ of the network such that $\textbf{P}^{-1}_{\tau} = W_{\tau} \otimes I_3,$ the information matrix for relative measurement between robots can be written as  
\begin{equation} \label{eqn:team_relative_meas_info_mat}
    \hat{\textbf{H}}_{\tau} =  L_{\tau} \otimes I_{3 },
\end{equation}
where $L_{\tau} = C_{\tau}W_{\tau}C_{\tau}^T$ is the Laplacian matrix of the network at time $\tau$. 

The $M-$step measurement model is given by 
\begin{equation}    \label{eqn:team_relative_meas_T_step}
    \bm{\xi}_{t:t+M} = \mathcal{C}_{t:t+M} \textbf{x}_{t:t+M} + \bm{\zeta}_{t:t+M},
\end{equation}
where $\mathcal{C}_{t:t+M} = \text{blkdiag}\left[(C_{t} \otimes I_{3})^T, \dots, (C_{t+M} \otimes I_{3})^T\right]$ is block diagonal augmentation of matrices $(C_{\tau} \otimes I_{3})^T $ for all $\tau \in [t:t+M].$
The information matrix of $\textbf{x}_{t:t+M}$ obtained using relative measurements can be written using \eqref{eqn:team_relative_meas_T_step} as 
\begin{equation}\label{eqn:team_relative_meas_info_mat_t_M}
   \hat{\textbf{H}}_{t:t+M} = \mathcal{L}_{t:t+M},
\end{equation}
where $\mathcal{L}_{t:t+M} = \text{blkdiag}\left[(L_{t} \otimes I_{3}), \dots, (L_{t+M} \otimes I_{3})\right]$ is block diagonal augmentation of matrices $(L_{\tau} \otimes I_{3}) $ for all $\tau \in [t, t+M].$
The information matrix $\hat{\textbf{H}}_{t:t+M}$ quantifies the amount of information of $\textbf{x}_{t : t + M}$ over the time horizon $M$ due to the network $\mathcal{G}_{\tau}.$
The information matrix of $\textbf{x}_{t : t + M}$ using the relative measurements can be updated according to 
\begin{equation} \label{eqn:inform_matrix_update_rel_meas}
    \Tilde{\textbf{H}}_{t: t + M} = \bar{\textbf{H}}_{t: t + M} +  \hat{\textbf{H}}_{t:t+M}. 
\end{equation}
The information matrix $ \Tilde{\textbf{H}}_{t: t + M}$ contains information of $\textbf{x}_{t : t + M}$ available through system dynamics \eqref{eqn:robot_dynamics_team} and relative measurement model \eqref{eqn:team_relative_meas_T_step}.
The following subsection delves into the attention-based vision model proposed by \cite{carlone2019attention}. We will discuss how the information extracted from features identified by this model can be integrated into our position estimation process.

\subsection{Model for Vision System}

Let us denote the set of all features available at any time step by $\Theta_t$,  the rotation matrices describing the orientation of the $i$'th camera and robot by $R_c^i \in $ SO(3)  and $R_{\tau}^i \in$ SO(3) respectively, the position vector of the feature $f \in \Theta_t$ by $y_f \in \R^3$, the position vector of the $i$'th camera with respect to $i$'th robot by $x_c^i$, and the unit vector (with respect to $i$'th camera) corresponding to pixel measurement of feature $f \in \Theta_t$ at time $\tau$ by $(u_{\tau}^{i,f})^T \in \R^3.$
For a given vector $(u_{\tau}^{i,f})^T $, its cross product with another vector $v$ can be written as the product of a skew-symmetric matrix $ U_{\tau}^{i,f}$ and vector $v$ as 
\begin{align*}
    (u_{\tau}^{i,f})^T \times v =  U_{\tau}^{i,f} v.
\end{align*}
Then the noisy vision model for the $i$'th robot proposed in \cite{carlone2019attention} is given by 
\begin{equation}\label{eqn: noisy_vision_model_ith}
    U_{\tau}^{i,f}(R_{\tau}^i R_c^i)^T \left(y_f - ({x}_{\tau}^i + R_{\tau}^i x_c^i)  \right) = {\eta}_{\tau}^{i,f}. 
\end{equation}
The model in \eqref{eqn: noisy_vision_model_ith} can be written equivalently as
\begin{equation}\label{eqn: noisy_vision_model_ith_alt}
    {z}_{\tau}^{i,f} = U_{\tau}^{i,f}(R_{\tau}^i R_c^i)^T  \left({x}_{\tau}^i - y_f \right) + {\eta}_{\tau}^{i,f}, 
\end{equation}
where ${\eta}_{\tau,T}^{i,f}$ is the measurement noise such that ${\eta}_{\tau,T}^{i,f} \sim \mathcal{N}(0,\sigma_i^2 I_3)$ and $ {z}_{\tau}^{i,f} = (U_{\tau}^{i,f})^T (R_c^i)^Tx_c^i.$

The vision model for the team of robots can be written by stacking the model \eqref{eqn: noisy_vision_model_ith_alt} for all robots as 
\begin{equation}\label{eqn: vision_model_team}
    {z}_{\tau}^f = F_{\tau}^f\textbf{x}_{t : t + M}+ E_{\tau}^f y_f  + {\eta}_{\tau}^f, 
\end{equation}
for appropriate matrices $F_{\tau}^f$ and $E_{\tau}^f.$

We assume that robots are capable of running $M-$step forward simulations to determine the number of frames $n_f$ in which the feature $f \in \Theta_t$ is visible.
The overall $M-$step vision model for a given feature $f$ can be written by vertically stacking \eqref{eqn: vision_model_team} as 
\begin{equation}\label{eqn: camera_model_forward}
    \textbf{z}_{t: t + M}^f = \textbf{F}_{t : t + M}^{f} \textbf{x}_{t : t + M} + \textbf{E}_{t: t + M}^{f} y_f + \bm{\eta}_{t: t + M}^f.
\end{equation}
Let us denote the covariance matrix of $\bm{\eta}_{t:t+M}^f$ by 
\[\mathbb{E}\left[\bm{\eta}_{t: t+M}^f \left(\bm{\eta}_{t:t+M}^{f}\right)^T \right] = \sigma_i^2 I_{3n_f}\in S^{3n_f}_{++}.\]
The information matrix of the parameters $\textbf{x}_{t : t + M}$ and $y_f$ is given by 
\begin{equation} \label{eqn: Omega_mat_1}
    \Omega_{\star}^{f} = \sigma_i^{-2}
            \begin{bmatrix}
         \left(\textbf{F}_{\star}^{f}\right)^T\textbf{F}_{\star}^{f} & \left(\textbf{F}_{\star}^{f}\right)^T\textbf{E}_{\star}^{f}\\
         \left(\textbf{E}_{\star}^{f}\right)^T \textbf{F}_{\star}^{f} & \left(\textbf{E}_{\star}^{f}\right)^T\textbf{E}_{\star}^{f}
    \end{bmatrix},
\end{equation}
where $\star = t:t+M$ is introduced for simplicity of notation.
The information matrix corresponding to the parameters $\textbf{x}_{t : t + M}$ is obtained by taking the Schur complement of the block $\left(\textbf{E}_{\star}^{f}\right)^T\textbf{E}_{\star}^{f}$ and can be written as 
\begin{equation}\label{eqn:Hf_mat}
    \scalebox{0.85}{$
        \hspace{-0.3cm}\textup{\bf{H}}_{\star}^f = \sigma_i^{-2}\Bigg(\left(\textbf{F}_{\star}^{f}\right)^T\textbf{F}_{\star}^{f} - \left(\textbf{F}_{\star}^{f}\right)^T\textbf{E}_{\star}^{f}\left(\left(\textbf{E}_{\star}^{f}\right)^T \textbf{E}_{\star}^{f}\right)^{-1}\left(\textbf{E}_{\star}^{f}\right)^T\textbf{F}_{\star}^{f}\Bigg).
    $}
\end{equation}
The information matrix $\textup{\bf{H}}_{\star}^f = \textup{\bf{H}}_{t:t+M}^f$ specified in \eqref{eqn:Hf_mat} is a measure of amount of information contained by a feature $f$ for estimation of $\textbf{x}_{t : t + M}.$

For any new observations by tracking a feature $f$, the information matrix of $\textbf{x}_{t:t+M}$, after accounting for relative measurements is updated as 
\begin{equation} \label{eqn:info_matrix_update_feature_f}
    \textbf{H}_{\star} \left(\{f\}\right) = \Tilde{\textbf{H}}_{\star} + \textup{\bf{H}}_{\star}^f.
\end{equation}
Since each feature's contribution is independent, the updates to the corresponding information matrices for a set $\Phi_t$ of selected features can be done by adding the individual information matrices according to the following:
\begin{equation}\label{eqn:infor_mat_update_feature}
    \textbf{H}_{\star} \left(\Phi_t\right) = \Tilde{\textbf{H}}_{\star} + \sum_{f \in \Phi_t}^{}\textup{\bf{H}}_{\star}^f.
\end{equation}
After accounting for all visible features in the set $\Phi_t,$ the information matrix can be inverted to obtain the covariance matrix according to
\begin{equation}\label{eqn:convert_inform_mat_to_covar}
    \bm{\Sigma}_{\star}(\Phi_t) =  \textbf{H}_{\star} \left(\Phi_t\right)^{-1}.
\end{equation}
To select a feature, it must be triangulated and the information matrix $\left(\textbf{E}_{\star}^{f}\right)^T\textbf{E}_{\star}^{f}$ corresponding to its position vector $y_f$ must be invertible.

To identify the most informative features, we define suitable metrics that quantify information content or estimated accuracy. To evaluate the information content of features, several performance measures can be employed. We discuss some of these measures next.
\begin{table}[t]
    
    \centering
    
        \caption{Performance Measures}
        {\renewcommand{\arraystretch}{2}
    \begin{tabular}{lcc}
    \hline
    \hline
        Performance Measures & Matrix Operator Form \\
        \hline
        Variance of the Error $\rho_v(\textup{\textbf{H}}_{\star}(\Phi_t))$ & $\textup{Tr}\left(\textup{\textbf{H}}_{\star}\left(\Phi_t\right)^{-1}\right)$\\
        Differential Entropy $\rho_e(\textup{\textbf{H}}_{\star}(\Phi_t))$ &  $- \log\left(\textup{det}\left (\textup{\textbf{H}}_{\star}(\Phi_t)\right)\right )$\\
        Spectral Variance $\rho_{\lambda}(\textup{\textbf{H}}_{\star}(\Phi_t)) $ & $\lambda_{\textup{min}}\left(\textup{\textbf{H}}_{\star}(\Phi_t)^{-1}\right)$\\
        \hline
        \hline
    \end{tabular}}
    \label{tab: Performance Measures}
\end{table}

\subsection{Performance Measures}
We quantify the informativeness of a feature subset,  $\Phi_t \subset \Theta_t$, by employing established performance measures  \cite{hossein2020feature_fast, arash2022space_time_sampling, carlone2019attention}. These performance measures are monotonically increasing map of the covariance matrix, which is defined in \eqref{eqn:convert_inform_mat_to_covar}. For accurate robot team localization, lower values of these performance measures indicate a better estimate of the robots' positions. Specific examples of these performance measures are listed in Table \ref{tab: Performance Measures}.

As robots share relative measurement information to localize themselves, the graph Laplacian $L_{\tau}$, which describes their communication network, plays a crucial role in determining the information matrix of their positions. 
In the next result, we show the monotonicity of the performance measures as a function of the graph Laplacian. 
\begin{theorem} \label{thm:performance_measure_and_connectivity}
The performance measures stated in Table \ref{tab: Performance Measures} are monotonically decreasing as connectivity of the communication graph $\mathcal{G}_{\tau}$ is increased, i.e., if $\mathcal{L}_{t:t+M} \preceq \mathcal{L'}_{t:t+M}$, then $\rho_{\square}(\textup{\textbf{H}}'_{\star}(\Phi_t))\leq  \rho_{\square}(\textup{\textbf{H}}_{\star}(\Phi_t))$ in which $\square \in \{v,e,\lambda\}$, $\textup{\textbf{H}}'_{\star}(\Phi_t))$ and $\textup{\textbf{H}}_{\star}(\Phi_t))$ are information matrices corresponding to $\mathcal{L'}_{t:t+M}$ and $\mathcal{L}_{t:t+M}$ respectively. 
\end{theorem}

Theorem \eqref{thm:performance_measure_and_connectivity} establishes that improved network connectivity among agents leads to better estimation quality for the robots' position vectors. This indicates that if the robots share relative information over a strongly connected communication graph, the uncertainty in estimate of their positions will decrease. This finding is crucial for designing effective communication graphs for the robot team. By enabling them to share relative information efficiently (strongly connected graphs), we can significantly reduce their position estimation uncertainty.


\section{Randomized Sampling Algorithm}\label{sec:random_algorithm}
This section presents a randomized algorithm for efficiently sampling a subset of candidate information matrices from the entire set.  Originally developed for graph sparsification \cite{spielman2011graph_eff_resistance}, this class of algorithms has found broader applicability in diverse fields, including network observability \cite{arash2022space_time_sampling}. Unlike the greedy algorithm, which can be computationally expensive, the randomized algorithm achieves similar results with lower computational complexity \cite{hossein2020feature_fast}. To quantify the level of uncertainty in our estimates, we further analyze the algorithm's performance by establishing probabilistic bounds for estimation accuracy.

We define the maximal information matrix as the sum of information matrices of all the available features $f \in \Theta_t$ such  that 
\begin{equation}\label{eqn:max_info_mat}
    {\textbf{H}_{\star}}(\Theta_t) = \Tilde{\textbf{H}}_{\star} + \sum_{f\in \Theta_t}^{} \textbf{H}_{\star}^f.
\end{equation}
The maximal matrix specified in \eqref{eqn:max_info_mat} can be decomposed into a sum of $|\Theta_t|$ individual matrices as 
\begin{equation}\label{eqn:indiv_info_mat}
   \Tilde{\textbf{H}}_{\star}^f:= \dfrac{1}{|\Theta_t|} \Tilde{\textbf{H}}_{\star} + {{\textbf{H}_{\star}^f}},
\end{equation}
for every $f \in \Theta_t$. Each of these matrices $ \Tilde{\textbf{H}}_{\star}^f$ captures the information content specific to a particular landmark observed by the robot team. In the following subsection, we  utilize the results from \eqref{eqn:max_info_mat} and \eqref{eqn:indiv_info_mat} to quantify the leverage score of each features $f \in \Theta_t$. 

\subsection{Sampling using Leverage Score}

Our leverage score-based sampling approach draws inspiration from the work by \cite{spielman2011graph_eff_resistance}, where leverage score is introduced as a measure of effective resistance in graphs. While this sampling approach is previously employed for information matrix selection \cite{hossein2020feature_fast}, our approach focuses on achieving tighter probabilistic guarantees for the algorithm. This is formally established in Theorem \eqref{thm:randmized_algorithm_probability_bound}.  Algorithm \ref{alg:sampl_leverage_score} outlines the steps involved in sampling a feature $f \in \Theta_t$ based on the leverage score of its information matrix.     

\begin{definition} \label{def:lever_score}
    For a feature $f \in \Theta_t$, the leverage score $r_f \in \R_+$ is defined as
    \begin{equation}\label{eqn:lever_score}
        r_f:= \textup{Tr} \left({\textup{\textbf{H}}_{\star}}(\Theta_t)^{-1} \Tilde{\textup{\textbf{H}}}_{\star}^f \right).
    \end{equation}
\end{definition}
The leverage score, denoted by $r_f$ for a feature $f \in \Theta_t$, quantifies the relative importance of each feature in the estimation process. This importance can be formalized by introducing a discrete probability measure over the set of all features, where features with higher leverage scores correspond to a higher probability of being selected during sampling.

\begin{remark}\label{rem: ind_prob}
Using the definition of leverage score in (\ref{def:lever_score}), we introduce a probability mass function (PMF) over the set of all available features, denoted by $\pi: \Theta_t \rightarrow [0, 1]$ such that
\begin{equation}\label{eqn:ind_prob}
    \pi_f:= \dfrac{r_f}{n},
\end{equation}
where $n = 3N(M+1)$, $N$ is the number of agents and $M$ is the prediction horizon. 
\end{remark} 

\begin{algorithm}[t]
\caption{Randomized Sampling Algorithm}\label{alg:sampl_leverage_score}
 \textbf{input}: initial information matrix $\Tilde{\textbf{H}}_{\star} = \bar{\textbf{H}}_{\star} + \hat{\textbf{H}}_{\star}$,\\
 \hspace*{0.5cm}set of all features $\Theta_t$, \\ 
\hspace*{0.5cm}and number of feature samples $q$\\
\textbf{output}: set of selected features $\Phi_t$, information matrix ${\textbf{H}_{\star}}$ \\ 
\textbf{initialize}: $\Phi_t = \emptyset$,  ${\textbf{H}_{\star}} = \Tilde{\textbf{H}}_{\star}$ \\
\textbf{for} k = 1 to $q$ \textbf{do}\\
\hspace*{0.5cm}sample a feature from $\Phi_t$ using distribution $\pi \rightarrow f$\\
\hspace*{0.5cm}select the information matrix 
\begin{align*}
    \textbf{H} \leftarrow {\textbf{H}_{\star}^f}
\end{align*}
\hspace*{0.5cm} \textbf{if} $f \notin \Phi_t$, \textbf{then}\\
\hspace*{0.75cm} add $f$ to $\Phi_t$\\
\hspace*{0.75cm} update the information matrix:
\begin{align*}
   \textbf{H}_{\star} \leftarrow \textbf{H} + {\textbf{H}_{\star}}
\end{align*}
\hspace*{0.5cm} \textbf{end if}\\
 \textbf{end for}
\end{algorithm}

\subsection{Performance Bound}

To evaluate how well our algorithm selects informative features, we compare the information captured by the selected features, which is quantified by $\textup{\textbf{H}}_{\star}(\Phi_t)$ to the information available from all possible features quantified by $\textup{\textbf{H}}_{\star}(\Theta_t)$. Theorem \ref{thm:randmized_algorithm_probability_bound} establishes a multiplicative bound on the information content of our algorithm. This bound holds with high probability, close to 3/4.

For the exposition of our next result, we define 
\begin{equation}\label{eqn:alpha}
    \kappa \in \{(0,3/4)~ | ~\log\left(n/{\kappa}\right)= O(\log n)\}.
\end{equation}

\begin{theorem}\label{thm:randmized_algorithm_probability_bound}
      For a given $\epsilon \in (0,1)$ and $\delta \in [\alpha,\frac{3}{4})$, suppose that the number of selected features $q = {2 n}\left(\log\left(n/{\delta}\right)\right)/\epsilon^2= O(n \log n /\epsilon^2) < |\Theta_t|$, then the information matrix of the set of features $\Phi_t \subset \Theta_t$ satisfies
        \begin{equation}\label{eqn:info_mat_cone_inequality}
        \textup{\textbf{H}}_{\star}(\Phi_t) \succeq \frac{(1-\epsilon)}{4\bar{\chi}} \textup{\textbf{H}}_{\star}(\Theta_t),
    \end{equation}
     with probability greater than $\frac{3}{4} -\delta$ for a number $\bar{\chi}$.
\end{theorem}
The cone inequality obtained in Theorem \ref{thm:randmized_algorithm_probability_bound} helps us to quantify bounds on the accuracy of the estimated positions of robots using performance measures listed in Table \ref{tab: Performance Measures}. Since performance measures are monotonically decreasing map of $\textup{\textbf{H}}_{\star}(\Phi_t)$, a probabilistic bound follows trivially.

Building upon Theorem \ref{thm:performance_measure_and_connectivity}, which established the link between network connectivity and performance measures, we now explore another key implication of the communication network structure on the feature selection problem. The following theorem focuses on the intricate relationship between the leverage score of a feature $f$ and the specific properties of the graphical network $\mathcal{G}_{\tau}$.

\begin{figure}
    \centering
    \input{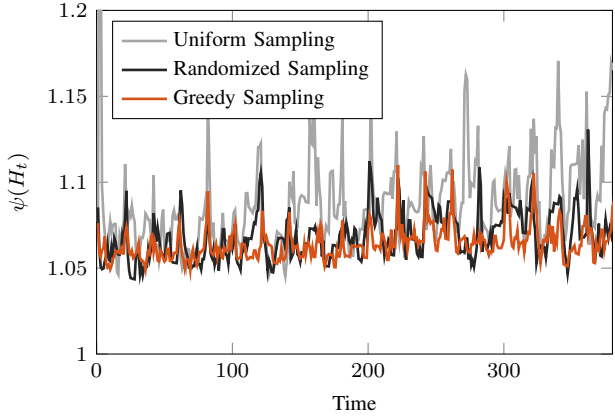}
    \caption{Ratio of spectral norm of estimated covariance to spectral norm of covariance when states are measured \eqref{eqn:estimation_covariance_ratio}.
    }
    \label{fig:covariance ratio}
\end{figure}

\begin{theorem}\label{thm:leverage_score_monotonicity}
For a given point feature $f \in \Theta_t$, the leverage score, as defined in \ref{eqn:lever_score}, is dependent on the underlying graph Laplacian of the communication network. It is monotonically decreasing when
\begin{equation}
    \mathbf{H}_{\star}^{f} \succeq \frac{1}{|\Theta_t|} \sum_{f \in \Theta_t} \mathbf{H}_{\star}^f,
\end{equation}
meaning that if $\mathcal{L}_{t:t+M} \preceq \mathcal{L}'_{t:t+M}$, then $r'_{f} \leq r_{f}$, where $r'_{f}$ and $r_{f}$ represent the leverage scores corresponding to the network information matrices $\mathcal{L}'_{t:t+M}$ and $\mathcal{L}_{t:t+M}$, respectively. In contrast, the leverage score is monotonically increasing under the condition
\begin{equation}
    \mathbf{H}_{\star}^{f} \preceq \frac{1}{|\Theta_t|} \sum_{f \in \Theta_t} \mathbf{H}_{\star}^f,
\end{equation}
which implies that if $\mathcal{L}_{t:t+M} \preceq \mathcal{L}'_{t:t+M}$, then $r_{f} \leq r'_{f}$.
\end{theorem}

Theorem \ref{thm:leverage_score_monotonicity} states that as the connectivity of the underlying communication graph of the network $\mathcal{G}_{\tau}$ increases, the leverage score of all features tend towards the mean leverage score. This has significant implications. In a strongly connected network of agents, all features become increasingly similar in their informativeness (leverage score). This allows for uniform random sampling of features to achieve good estimation quality. Consequently, the computational complexity of the algorithm is significantly reduced. Calculating the leverage score for features is no longer necessary, as random sampling becomes equally effective.

In the next section, we thoroughly evaluate the performance of the algorithm \ref{alg:sampl_leverage_score}.


\section{Case Study} \label{sec: case_study}
We conduct simulation studies to evaluate the performance of the randomized algorithm \ref{alg:sampl_leverage_score} against greedy and uniform sampling algorithms. To establish the context of our simulations, we begin by detailing the robot models and the features of the environment.

\begin{figure}[t]
    \centering
    \input{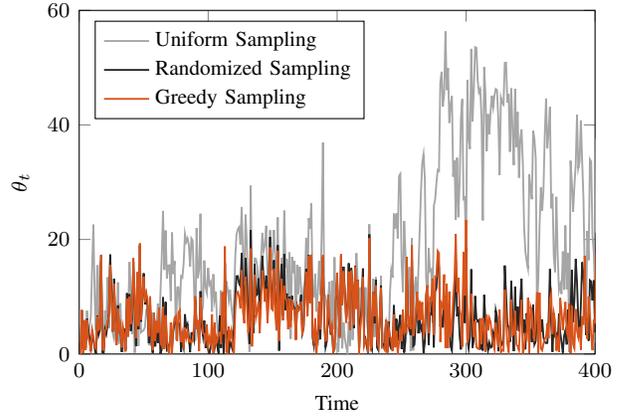}
    \caption{Multi-agent localization error for different feature selection algorithms.}
    \label{fig:estimation_error}
\end{figure}

\begin{figure*}[t]
    \centering
    \input{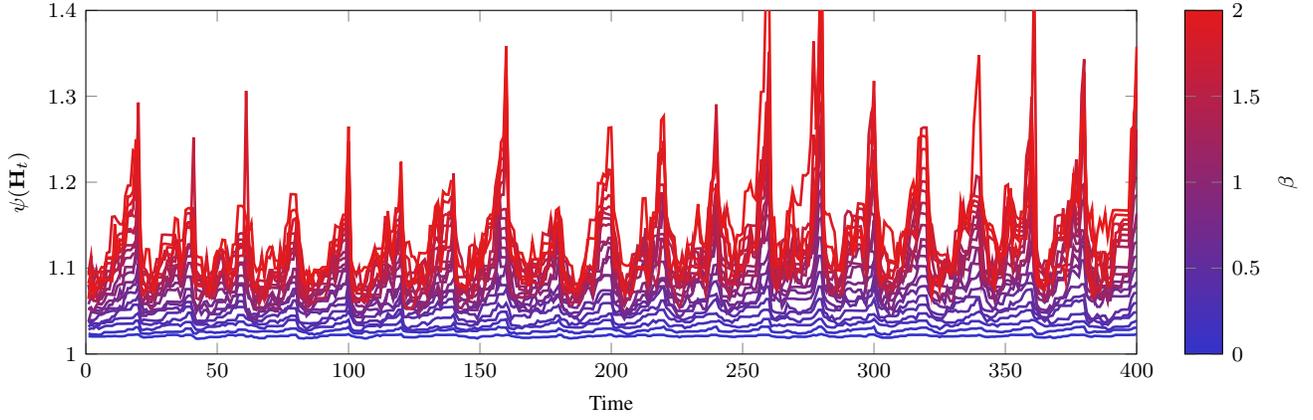}
    \caption{Impact of network connectivity on estimation covariance.} 
    \label{fig:network_connectivity_vs covariance}
\end{figure*}

\subsection{Simulation Setup: Robots and Environment}
We consider a team of $N = 10$ robots that can share relative measurements with each other over a communication graph denoted as $\mathcal{G}{\tau}$. The weights of the edges in $\mathcal{G}{\tau}$ represent the communication strength between robots. These weights decay exponentially with the increasing distance between robots $i$ and $j$ as follows:
\begin{equation}\label{eqn:exponential_weight_decay}
    \omega_{ij}(\tau) = \alpha e^{-\beta \|x_{\tau}^i - x_{\tau}^j\|},
\end{equation}
where $\omega_{ij}(\tau)$ represents the weight of the communication edge between robots $i$ and $j$ for all $i,j \in \{1, \dots, N\}$, $\alpha$ and $\beta$ are constants that control the decay rate.  Essentially, robots closer together have stronger communication links (higher weight), while communication weakens as the distance increases. This approach mirrors real-world limitations in robot communication range or signal strength. Furthermore, this weight change is linked to the covariance matrix of the relative measurement model, implying that the uncertainty in robot localization grows as the distance between communicating robots increases. By analyzing these weight changes and their impact on the estimation covariance matrix, we can gain valuable insights into how the network topology influences the overall accuracy and performance of robot localization algorithms.

With position vector $x_{\tau} = [p_{\tau},y_{\tau},z_{\tau}]^T$ \footnote{We use letter $p$ for the first coordinate to prevent conflict with the use of position vector $x$.}, the dynamics of the $i$'th robot is given by 
\begin{equation}\label{eqn:robot_dynamics_simulation}
    \begin{aligned}
        \begin{cases}
            p_{{\tau}+1}^{i} &= p_{\tau}^{i} + u_{\tau}^{p,i} + \delta^{p,i}_{\tau}\\
            y_{{\tau}+1}^{i} &= y_{\tau}^{i} + u_{\tau}^{y,i} + \delta^{y,i}_{\tau}\\
            z_{{\tau}+1}^{i} &= z_{\tau}^{i} + u_{\tau}^{z,i} + \delta^{z,i}_{\tau}
        \end{cases},
    \end{aligned}
\end{equation}
where $u_{\tau} = [u_{\tau}^{p,i},u_{\tau}^{y,i},u_{\tau}^{z,i} ]^T$ is the control input of the $i$'th robot and $\delta_{\tau}^{i} = [\delta_{\tau}^{p,i},\delta_{\tau}^{y,i},\delta_{\tau}^{z,i}]^T$ is the random process describing the uncertainty propagation due to the motion of the $i$'th robot.

We assume that each robot navigates along a pre-defined trajectory. The trajectory of the $i$'th robot is given by
\begin{equation}\label{eqn:robot_trajectory_simulation}
    \begin{aligned}
                \begin{cases}
            p_{{\tau}}^{i,\textup{ref}} &= 10^4\left(\left(i+\frac{1}{2}\right)  + 0.1 \sin\left(\frac{\pi}{10}\tau\right)\right) \\
            y_{{\tau}}^{i,\textup{ref}} &= -2 \left(\frac{t_e+M}{t_e}\right)+10\frac{\tau}{t_e}\\
            z_{{\tau}}^{i,\textup{ref}} &= \sin\left(\frac{4\pi}{N t_e}\left(i - \frac{N}{2}\right)\tau\right)
        \end{cases},
    \end{aligned}
\end{equation}
where $t_e = 200$, $N=10$ is the number of robots and $M=20$ is the prediction horizon.
Additionally, we assume that the Euler angles, which describe the absolute orientation of the $i$'th camera at any given time $\tau$ are given by 
\begin{equation}\label{eqn:Euler_angles}
    \begin{aligned}
        \begin{cases}
            \alpha_{\tau}^{i} = 0\\
            \beta_{\tau}^{i} = \frac{\pi}{2}+ \frac{\pi}{5} \sin\left(\frac{4\pi}{N t_e}\left(i - \frac{N}{2}\right)\tau\right)\\
            \gamma_{\tau}^{i} = \frac{\pi}{2} + \frac{\pi}{10} \sin\left(\frac{\pi}{N}\left(i - \frac{N}{2}\right) + \frac{\pi}{10} \tau\right)
        \end{cases},
    \end{aligned}
\end{equation}
where the sequence of rotations is $z-y-p.$

Figure~\ref{fig:Robots_environment} illustrates the environment from different perspectives, including snapshots and the trajectories of the robots over time.

\subsection{Feature Selection Algorithms and Comparison Metrics}
We compare the proposed Randomized sampling algorithm with Uniform randomized sampling and the Greedy sampling algorithm as the baseline algorithms. In randomized sampling, we select features based on a probability distribution derived from the leverage score, outlined in Algorithm~\ref{alg:sampl_leverage_score}.  We assign a uniform probability to all features instead of using the leverage score for uniform randomized sampling. For the greedy algorithm, we iteratively select a single feature at a time that leads to the most improvement in a chosen performance measure.

To compare the performance of different algorithms, we employ the mean square estimation error to measure how close are the state estimation to the true trajectory. The mean square estimation error is denoted by
\begin{equation}\label{eqn:estimation_error}
    \theta_t = {\|\bm{x}_{t}- \bm{\mu}_{t}\|_2}.
\end{equation}
To measure which algorithm has a better performance in uncertainty estimation, we employ the ratio of the spectral norm of estimated covariance to the spectral norm of covariance when states are measured. This ratio is given by
\begin{equation}\label{eqn:estimation_covariance_ratio}
    \psi(H_t) =  \dfrac{ \rho_\lambda ({H}_{t})}{  \lambda_{\min}( I_N \otimes \Lambda_t )}, 
\end{equation}
 a lower ratio suggests that feature selection algorithm has effectively reduced the uncertainty in robots' position.

\subsection{Simulation Results}
We present a comparative analysis of several algorithms through extensive simulations. By analyzing diverse scenarios, we gain valuable insights into the strengths and weaknesses of each algorithm.
Figure~\ref{fig:estimation_error} depicts the mean squared estimation error for all three feature selection algorithms. Our proposed algorithm achieves comparable performance to the greedy algorithm, while both outperform the uniform sampling approach by a significant margin. Figure~\ref{fig:covariance ratio} illustrates the reduction in localization uncertainty, quantified by the norm of estimated covariance, achieved by the different feature selection algorithms. Our proposed algorithm demonstrates a similar level of uncertainty reduction compared to the greedy algorithm. Notably, both approaches significantly outperform uniform sampling. 

While our randomized algorithm achieves slightly lower accuracy compared to the greedy approach as evident in Figures \ref{fig:estimation_error} and \ref{fig:covariance ratio}, it offers significant advantages in terms of computational complexity. These results underscore the critical role of feature selection algorithms in achieving accurate multi-agent localization. By selecting informative features, we significantly improve the quality of localization.

\begin{figure}[t!]
    \centering
    \begin{subfigure}[b]{0.24\textwidth}
        \centering
%
%
\definecolor{mycolor1}{rgb}{0.00000,0.44700,0.74100}%

\begin{tikzpicture}
\pgfplotsset{scaled x ticks=false}
\tikzstyle{every node}=[font=\footnotesize]
\begin{axis}[%
width=1.2in,
height=1.2in,
scale only axis,
xmin=0,
xmax=0.06,
yticklabel style={%
                 /pgf/number format/.cd,
                     fixed,
                     fixed zerofill,
                     precision=2,
                     },
xticklabel style={%
                 /pgf/number format/.cd,
                     fixed,
                     fixed zerofill,
                     precision=2,
                     },
ymin=0,
ymax=0.15,
ylabel={Frequency},
xticklabels={$0$,$0$,$0.02$,$0.04$,$0.06$},
legend style={legend plot pos=none},
]
\addplot[ybar interval, fill=mycolor1, fill opacity=0.3, draw=white!60!black, area legend] table[row sep=crcr] {%
x	y\\
0	0\\
0.001	0\\
0.002	0\\
0.003	0\\
0.004	0\\
0.005	0\\
0.006	0\\
0.007	0\\
0.008	0.00280112044817927\\
0.009	0.00360144057623049\\
0.01	0.0220088035214086\\
0.011	0.0696278511404562\\
0.012	0.105642256902761\\
0.013	0.114445778311325\\
0.014	0.108843537414966\\
0.015	0.106442577030812\\
0.016	0.10844337735094\\
0.017	0.0820328131252501\\
0.018	0.0796318527410964\\
0.019	0.0716286514605842\\
0.02	0.0492196878751501\\
0.021	0.0312124849939976\\
0.022	0.0196078431372549\\
0.023	0.0120048019207683\\
0.024	0.00640256102440976\\
0.025	0.00440176070428171\\
0.026	0.00040016006402561\\
0.027	0.00160064025610244\\
0.028	0\\
0.029	0\\
0.03	0\\
0.031	0\\
0.032	0\\
0.033	0\\
0.034	0\\
0.035	0\\
0.036	0\\
0.037	0\\
0.038	0\\
0.039	0\\
0.04	0\\
0.041	0\\
0.042	0\\
0.043	0\\
0.044	0\\
0.045	0\\
0.046	0\\
0.047	0\\
0.048	0\\
0.049	0\\
0.05	0\\
0.051	0\\
0.052	0\\
0.053	0\\
0.054	0\\
0.055	0\\
0.056	0\\
0.057	0\\
0.058	0\\
0.059	0\\
0.06	0\\
};
\addlegendentry{$\beta = 0$}

\end{axis}
\end{tikzpicture}%
    \end{subfigure}%
    \hfill
    \begin{subfigure}[b]{0.24\textwidth}
        \centering
%
%
\definecolor{mycolor1}{rgb}{0.00000,0.44700,0.74100}%

\begin{tikzpicture}
\pgfplotsset{scaled x ticks=false}
\tikzstyle{every node}=[font=\footnotesize]
\begin{axis}[%
width=1.2in,
height=1.2in,
scale only axis,
xmin=0,
xmax=0.06,
yticklabel style={%
                 /pgf/number format/.cd,
                     fixed,
                     fixed zerofill,
                     precision=2,
                     },
xticklabel style={%
                 /pgf/number format/.cd,
                     fixed,
                     fixed zerofill,
                     precision=2,
                     },
ymin=0,
ymax=0.15,
xticklabels={$0$,$0$,$0.02$,$0.04$,$0.06$},
legend style={legend plot pos=none},
ytick=\empty,
]
\addplot[ybar interval, fill=mycolor1, fill opacity=0.3, draw=white!60!black, area legend] table[row sep=crcr] {%
x	y\\
0	0\\
0.001	0\\
0.002	0\\
0.003	0\\
0.004	0.00320128051220488\\
0.005	0.0104041616646659\\
0.006	0.0448179271708683\\
0.007	0.0656262505002001\\
0.008	0.056422569027611\\
0.009	0.0664265706282513\\
0.01	0.0644257703081232\\
0.011	0.0620248099239696\\
0.012	0.040016006402561\\
0.013	0.0520208083233293\\
0.014	0.0444177671068427\\
0.015	0.0428171268507403\\
0.016	0.0416166466586635\\
0.017	0.0364145658263305\\
0.018	0.0368147258903561\\
0.019	0.0436174469787915\\
0.02	0.0404161664665866\\
0.021	0.0372148859543818\\
0.022	0.0332132853141257\\
0.023	0.0332132853141257\\
0.024	0.0212084833933573\\
0.025	0.0256102440976391\\
0.026	0.0196078431372549\\
0.027	0.0148059223689476\\
0.028	0.0136054421768707\\
0.029	0.0140056022408964\\
0.03	0.00760304121648659\\
0.031	0.00440176070428171\\
0.032	0.00560224089635854\\
0.033	0.0040016006402561\\
0.034	0.00320128051220488\\
0.035	0.00200080032012805\\
0.036	0.00160064025610244\\
0.037	0.00200080032012805\\
0.038	0.00040016006402561\\
0.039	0.00240096038415366\\
0.04	0.00040016006402561\\
0.041	0.00040016006402561\\
0.042	0.00040016006402561\\
0.043	0.000800320128051221\\
0.044	0.000800320128051221\\
0.045	0\\
0.046	0\\
0.047	0\\
0.048	0\\
0.049	0\\
0.05	0\\
0.051	0\\
0.052	0\\
0.053	0\\
0.054	0\\
0.055	0\\
0.056	0\\
0.057	0\\
0.058	0\\
0.059	0\\
0.06	0\\
};
\addlegendentry{$\beta=1$}

\end{axis}
\end{tikzpicture}%
    \end{subfigure}%
    \hfill
    \begin{subfigure}[b]{0.24\textwidth}
        \centering
%
%
\definecolor{mycolor1}{rgb}{0.00000,0.44700,0.74100}%
\begin{tikzpicture}
\pgfplotsset{scaled x ticks=false}
\tikzstyle{every node}=[font=\footnotesize]
\begin{axis}[%
width=1.2in,
height=1.2in,
scale only axis,
xmin=0,
xmax=0.06,
yticklabel style={%
                 /pgf/number format/.cd,
                     fixed,
                     fixed zerofill,
                     precision=2,
                     },
xticklabel style={%
                 /pgf/number format/.cd,
                     fixed,
                     fixed zerofill,
                     precision=2,
                     },
ymin=0,
ymax=0.15,
xlabel={Leverage score},
ylabel={Frequency},
xticklabels={$0$,$0$,$0.02$,$0.04$,$0.06$},
legend style={legend plot pos=none},
]
\addplot[ybar interval, fill=mycolor1, fill opacity=0.3, draw=white!60!black, area legend] table[row sep=crcr] {%
x	y\\
0	0\\
0.001	0\\
0.002	0.00240096038415366\\
0.003	0.0040016006402561\\
0.004	0.0352140856342537\\
0.005	0.0580232092837135\\
0.006	0.0532212885154062\\
0.007	0.0556222488995598\\
0.008	0.0600240096038415\\
0.009	0.0520208083233293\\
0.01	0.054421768707483\\
0.011	0.0328131252501\\
0.012	0.0368147258903561\\
0.013	0.0388155262104842\\
0.014	0.0368147258903561\\
0.015	0.0364145658263305\\
0.016	0.0344137655062025\\
0.017	0.030812324929972\\
0.018	0.0352140856342537\\
0.019	0.0320128051220488\\
0.02	0.038015206082433\\
0.021	0.0264105642256903\\
0.022	0.0356142456982793\\
0.023	0.0256102440976391\\
0.024	0.0248099239695878\\
0.025	0.0240096038415366\\
0.026	0.0172068827531012\\
0.027	0.021608643457383\\
0.028	0.0120048019207683\\
0.029	0.0120048019207683\\
0.03	0.0112044817927171\\
0.031	0.0108043217286915\\
0.032	0.0112044817927171\\
0.033	0.00760304121648659\\
0.034	0.00320128051220488\\
0.035	0.00440176070428171\\
0.036	0.00520208083233293\\
0.037	0.00280112044817927\\
0.038	0.00200080032012805\\
0.039	0.00280112044817927\\
0.04	0.00160064025610244\\
0.041	0.00120048019207683\\
0.042	0.00320128051220488\\
0.043	0\\
0.044	0.00120048019207683\\
0.045	0.000800320128051221\\
0.046	0.000800320128051221\\
0.047	0.000800320128051221\\
0.048	0.00040016006402561\\
0.049	0.00040016006402561\\
0.05	0.00040016006402561\\
0.051	0.000800320128051221\\
0.052	0\\
0.053	0\\
0.054	0.000800320128051221\\
0.055	0\\
0.056	0\\
0.057	0\\
0.058	0\\
0.059	0\\
0.06	0\\
};
\addlegendentry{$\beta=2$}

\end{axis}
\end{tikzpicture}%
    \end{subfigure}%
    \hfill
    \begin{subfigure}[b]{0.24\textwidth}
        \centering
%
%
\definecolor{mycolor1}{rgb}{0.00000,0.44700,0.74100}%
\begin{tikzpicture}
\pgfplotsset{scaled x ticks=false}
\tikzstyle{every node}=[font=\footnotesize]
\begin{axis}[%
width=1.2in,
height=1.2in,
scale only axis,
xmin=0,
xmax=0.06,
yticklabel style={%
                 /pgf/number format/.cd,
                     fixed,
                     fixed zerofill,
                     precision=2,
                     },
xticklabel style={%
                 /pgf/number format/.cd,
                     fixed,
                     fixed zerofill,
                     precision=2,
                     },
ymin=0,
ymax=0.15,
xticklabels={$0$,$0$,$0.02$,$0.04$,$0.06$},
xlabel={Leverage score},
legend style={legend plot pos=none},
ytick=\empty
]
\addplot[ybar interval, fill=mycolor1, fill opacity=0.3, draw=white!60!black, area legend] table[row sep=crcr] {%
x	y\\
0	0\\
0.001	0\\
0.002	0.00640256102440976\\
0.003	0.0404161664665866\\
0.004	0.0608243297318928\\
0.005	0.0492196878751501\\
0.006	0.0512204881952781\\
0.007	0.0588235294117647\\
0.008	0.0496198479391757\\
0.009	0.0464185674269708\\
0.01	0.0352140856342537\\
0.011	0.0312124849939976\\
0.012	0.0280112044817927\\
0.013	0.0292116846738695\\
0.014	0.0392156862745098\\
0.015	0.0352140856342537\\
0.016	0.0320128051220488\\
0.017	0.0256102440976391\\
0.018	0.030812324929972\\
0.019	0.0304121648659464\\
0.02	0.0332132853141257\\
0.021	0.0272108843537415\\
0.022	0.0292116846738695\\
0.023	0.0256102440976391\\
0.024	0.0192076830732293\\
0.025	0.0220088035214086\\
0.026	0.0224089635854342\\
0.027	0.0176070428171269\\
0.028	0.014405762304922\\
0.029	0.014405762304922\\
0.03	0.0112044817927171\\
0.031	0.0108043217286915\\
0.032	0.00960384153661465\\
0.033	0.0100040016006403\\
0.034	0.00880352140856343\\
0.035	0.00720288115246098\\
0.036	0.00480192076830732\\
0.037	0.00520208083233293\\
0.038	0.00200080032012805\\
0.039	0.00600240096038415\\
0.04	0.00240096038415366\\
0.041	0.00200080032012805\\
0.042	0.00160064025610244\\
0.043	0.00160064025610244\\
0.044	0.00120048019207683\\
0.045	0.000800320128051221\\
0.046	0.00200080032012805\\
0.047	0.00160064025610244\\
0.048	0.00120048019207683\\
0.049	0.000800320128051221\\
0.05	0.00040016006402561\\
0.051	0.00040016006402561\\
0.052	0\\
0.053	0.000800320128051221\\
0.054	0\\
0.055	0.00040016006402561\\
0.056	0.000800320128051221\\
0.057	0\\
0.058	0.00040016006402561\\
0.059	0\\
0.06	0\\
};
\addlegendentry{$\beta=5$}

\end{axis}
\end{tikzpicture}%
    \end{subfigure}
    \caption{Histogram of leverage scores of features for varying communication strength $\beta$ and $\alpha=1$, where $\alpha$ and $\beta$ are as defined in \eqref{eqn:exponential_weight_decay}. 
    }
    \label{fig:histogram_of_probability}
\end{figure}
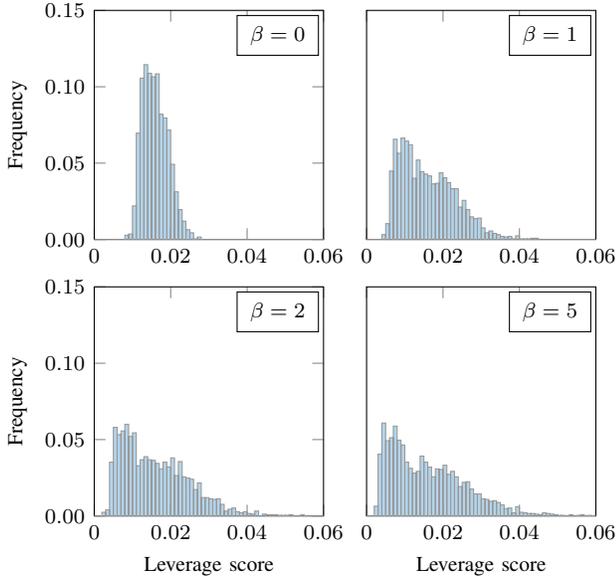

We now delve into the impact of network connectivity on multi-agent localization performance. Through extensive simulations, we examine how communication strength among robots influences the effectiveness of feature selection algorithms, ultimately affecting the accuracy of their position estimates. Figure~\ref{fig:network_connectivity_vs covariance} illustrates the crucial role network connectivity plays in reducing estimation uncertainty for multi-agent localization. As the connections between robots strengthen, quantified by a decrease in the parameter $\beta$ \eqref{eqn:exponential_weight_decay}, the covariance, a measure of uncertainty in their position estimates, noticeably diminishes. This observation aligns perfectly with our theoretical findings in Theorem~\ref{thm:performance_measure_and_connectivity}, which states that stronger network connectivity leads to reduced estimation error. Stronger communication links allow robots to share information more effectively, reducing the overall uncertainty in their localization task. The peaks observed in Figure~\ref{fig:network_connectivity_vs covariance} suggest that selected features are informative only within a specific prediction horizon. Figure~\ref{fig:histogram_of_probability} shows a histogram of feature leverage scores for varying communication strength $\beta$ as defined in \eqref{eqn:exponential_weight_decay}. As network connectivity strengthens (lower $\beta$), the distribution of feature leverage scores becomes more uniform. This is reflected in the steeper histogram for $\beta=0.$ This validates our result in Theorem~\ref{thm:leverage_score_monotonicity}, which states that leverage scores of features become more uniform with stronger network connectivity.


\section{Conclusion}

We propose a randomized algorithm for sparse visual feature selection, enabling efficient multi-agent localization.
Compared to uniform random sampling, our approach achieves significantly better performance by prioritizing informative features. While exhibiting comparable accuracy to the greedy algorithm, our randomized algorithm offers potential advantages in terms of improving algorithmic complexity. We also investigate the impact of network structure on multi-agent localization performance as higher connectivity translates to lower uncertainty in localization. Through extensive simulations and theoretical proofs, we demonstrate that in a strongly connected communication graph, the importance of features becomes more uniform. This allows for uniform random sampling, which significantly reduces the computational complexity. These findings suggest the promise of randomized algorithms for achieving efficient and accurate multi-agent localization with sparse feature sets.


\printbibliography

@article{spielman2011graph_eff_resistance,
  title={Graph Sparsification by Effective Resistances},
  author={Spielman,Daniel A and Srivastava, N.},
  journal={SIAM Journal of Computing},
  volume={40},
  number={6},
  pages={1913--1926},
  year={2011},
  publisher={}
}

@article{tropp2012tailbounds_random_matrix,
  title={User-Friendly Tail Bounds for Sums of Random Matrices},
  author={Tropp, Joel A},
  journal={Foundations of Computational Mathematics},
  volume={12},
  number={4},
  pages={389--434},
  year={2012},
  publisher={}
}

@article{thrun2004info_filter,
  title={Simultaneous Localization and Mapping with Sparse Extended Information Filters},
  author={Thrun, Sebastian A and Liu, Yufeng and Koller, Daphne and Ng, Andrew Y and Ghahramani, Zoubin and Durrant-Whyte, Hugh.},
  journal={The International Journal of Robotics Research},
  volume={23},
  number={7--8},
  pages={693--716},
  year={2004},
  publisher={}
}

@article{hossein2020feature_fast,
  title={Estimation with Fast Feature Selection in Robot Visual Navigation},
  author={Mousavi, Hossein K  and Motee, Nader},
  journal={IEEE Robotics and Automation Letters},
  volume={5},
  number={2},
  pages={3572--3579},
  year={2020}
}

@article{guadagnino2022fast_sparse_lidar,
  title={Fast Sparse LiDAR Odometry Using Self-Supervised Feature Selection on Intensity Images},
  author={Guadagnino, Tiziano  and Chen, Xieyuanli and Sodano, Matteo and Behley,Jens and Grisetti, Giorgio and Stachniss, Cyrill},
  journal={IEEE Robotics and Automation Letters},
  volume={7},
  number={3},
  pages={7597--7604},
  year={2022}
}

@article{arash2022space_time_sampling,
  title={Space Time Sampling for Network Observability},
  author={Amini, Arash and Mousavi, Hossein K and Sun, Qiyu and Motee, Nader},
  journal={IEEE Transactions on Control of Network Systems},
  volume={10},
  number={3},
  pages={1159--1171},
  year={2023}
}

@article{siami2018centrality_measures,
  title={Centrality Measures in Linear Consensus Networks with Structured Network Uncertainties},
  author={Siami, Milad and Bolouki, Sadegh and Bamieh, Bassam and Motee, Nader},
  journal={IEEE Transactions on Control of Network Systems},
  volume={5},
  number={3},
  pages={924--934},
  year={2018}
}

@article{carlone2019attention,
  title={Attention and Anticipation in Fast Visual-Inertial Navigation},
  author={Carlone, Luca  and Karaman, Sertac},
  journal={IEEE Transactions on Robotics},
  volume={35},
  number={1},
  pages={},
  year={2019}
}

@article{zhao2020good_feature_matching,
  title={Good Feature Matching: Towards Accurate, Robust VO/VSLAM With Low Latency},
  author={Zhao, Yipu  and Vela, Patricio A},
  journal={IEEE Transactions on Robotics},
  volume={36},
  number={3},
  pages={657--675},
  year={2020}
}

@article{sala2006landmark_selection_pose,
  title={Landmark Selection for Vision-based navigation},
  author={Sala, P and Sim, R and Shokoufandeh, A and Dickinson,S},
  journal={IEEE Transactions on Robotics},
  volume={22},
  number={2},
  pages={334--349},
  year={2006}
}

@article{lerner2007landmark_selection_task,
  title={Landmark Selection for Task-oriented navigation},
  author={Lerner, R and Rivlin, E and Shimshoni, I},
  journal={IEEE Transactions on Robotics},
  volume={23},
  number={3},
  pages={494--505},
  year={2007}
}

@article{gorbenko2012landmark_selection_minimalset,
  title={The Problem of Selection of a Minimal Set of Landmarks},
  author={Gorbenko, A and Popov, V},
  journal={Applied Mathematical Sciences},
  volume={6},
  number={95},
  pages={4729--4732},
  year={2012}
}

@article{Hossein2020characterization_performance,
  title={Explicit Characterization of Performance of a Class of Networked Linear Control Systems},
  author={Mousavi, Hossein K and Motee, Nader},
  journal={IEEE Transactions on Control of Network Systems},
  volume={7},
  number={4},
  pages={1688--1699},
  year={2020}
}

@article{nemhauser1978maximizing_submodular_Set_functions,
  title={An Analysis of Approximations for Maximizing Submodular Set Functions-I},
  author={Nemhauser, G.L.  and Wolsley, L.A. and Fischer, M.L.},
  journal={Mathematical Programming},
  volume={14},
  number={1},
  pages={40--53},
  year={1978}
}

@inproceedings{strasdat2009landmark_useful ,
  title={Which landmark is Useful? Learning Selection Policies for Navigation in Unknown Environments},
  author={Strasdat, H and Stachniss, C and Burgard, W},
  booktitle={IEEE International Conference on Robotics and Automation},
  pages={1410--1415},
  year={2009},
  organization={IEEE}
}

@inproceedings{mirzasoleiman2015stochastic_greedy ,
  title={Lazier Than Lazy Greedy },
  author={Mirzasoleiman, Baharan and Badanidiyuru, Ashwinkumar and Karbasi, Amin and Vondrak, Jan and Krause, Andreas},
  booktitle={Twenty-Ninth AAAI Conference on Artificial Inteliigence},
  pages={182--1818},
  year={2015},
  organization={}
}

@inproceedings{thrun2003info_filter_multi_robot_slam ,
  title={Multi-robot SLAM with Sparse Extended Information Filters},
  author={Thrun, Sebastian A and Liu, Yufeng},
  booktitle={Proceedings of the th International Symposium of Robotics Research (ISRR'03)},
  pages={},
  year={2003},
  organization={}
}

@inproceedings{jiao2022lidar_slam_feature_selection ,
  title={Greedy-Based Feature Selection for Efficient LiDAR SLAM},
  author={Jiao, Jianhao and Zhu, Yilong and Ye, Haoyang and Huang, Huaiyang and Yun, Peng and Jiang, Linxin and Wang, Lujia and Liu, Ming},
  booktitle={IEEE International Conference on Robotics and Automation},
  pages={5222--5228},
  year={2022},
  organization={IEEE}
}

@book{thrun2005_probabilistic_robotics,
    title = {{Probabilistic Robotics}},
    year = {2005},
    author = {Thrun, Sebastian and Burgard, Wolfram and Fox, Dieter},
    publisher = {MIT Press}
}

\appendix
\noindent \underline{\bf Proof of Theorem \ref{thm:performance_measure_and_connectivity}:}
 Using \eqref{eqn:team_relative_meas_info_mat_t_M}, \eqref{eqn:inform_matrix_update_rel_meas},and \eqref{eqn:info_matrix_update_feature_f}, one can show that
 \begin{equation*}
     \scalebox{1}{$
        \textbf{H}_{\star} \left(\Phi_t\right) =\Bar{\textbf{H}}_{\star} + \mathcal{L}_{t:t+M}+\sum_{f \in \Phi_t}^{}\textup{\bf{H}}_{\star}^f,
     $}
 \end{equation*}
and
\begin{equation*}
    \scalebox{1}{$
        \mathcal{L}_{t:t+M} \preceq \mathcal{L'}_{t:t+M} \implies  \textbf{H}'_{\star} \left(\Phi_t\right) \preceq  \textbf{H}_{\star} \left(\Phi_t\right).
    $}
\end{equation*}
Then, the result follows from the monotonicity of $\rho_{\square}(\cdot)$ as shown in Table \ref{tab: Performance Measures}.

\vspace{1mm}
\noindent \underline{\bf Proof of Theorem \ref{thm:randmized_algorithm_probability_bound}:}
The proof of this theorem leverages tail bounds for sums of random matrices. Specifically, we rely on Corollary 5.2 from \cite{tropp2012tailbounds_random_matrix} which provides guarantees on the concentration behavior of such sums. We state a version of the result here as Lemma \ref{lem:matrix_conc_ineq}.
\begin{lemma} \label{lem:matrix_conc_ineq}
    Let $X_1, \cdots, X_q$ be independent random $n-$dimensional positive semi-definite matrices such that $\|X_i\| \leq R$ almost surely for all $i \in \{1, \dots, q\}$, where $\|\cdot\|$ denotes the operator norm. 
    Consider $\Tilde{X} = \sum_{i = 1}^{q} X_i$ and let $\mu_{\textup{min}}$ be the minimum eigenvalue of expectation of $\Tilde{X}$, i.e.
\begin{equation*}
    \scalebox{1}{$
        \mu_{\textup{min}} := \lambda_{\textup{min}} \left(\mathbb{E}\left[\Tilde{X}\right]\right) \hspace{0.25cm} ,
    $}
\end{equation*}
then the following holds 
\begin{equation}    \label{eqn:tropp_matrix_chernoff_eq}
    \scalebox{0.9}{$
        \mathbb{P}\left[ \lambda_{\textup{min}} \left(\Tilde{X}\right)\leq (1-\epsilon) \mu_{\textup{min}}\right] \leq n  \left(\dfrac{e^{-\epsilon}}{\left(1 - \epsilon\right)^{1 - \epsilon}}\right)^{\mu_{\textup{min}}/R},
    $}
\end{equation}
for  $0< \epsilon < 1$. To obtain a tractable form of \eqref{eqn:tropp_matrix_chernoff_eq}, we use the following inequality, 
\begin{equation*}
    \scalebox{0.9}{$
        \left(\dfrac{e^{-\epsilon}}{\left(1 - \epsilon\right)^{1 - \epsilon}}\right) \leq e^{-\epsilon^2/2}, \hspace{0.5cm} \textup{for}~ 0< \epsilon < 1,
    $}
\end{equation*}
such that 
\begin{equation}\label{eqn:tropp_matrix_chernoff_eq_alt}
    \scalebox{0.9}{$
    \mathbb{P}\left[ \lambda_{\textup{min}} \left(\Tilde{X}\right)\geq (1-\epsilon) \mu_{\textup{min}}\right] \geq 1 - n e^{-\dfrac{\mu_{\textup{min}}\epsilon^2}{2R}}.
    $}
\end{equation}
\end{lemma}

Let us define the matrix
$\Bar{\textup{\textbf{B}}}_{\star}^{f} \in S^{n}_{++}$
for all $f \in \Theta_t$ such that 
\begin{equation} \label{eqn: B_mat}
    \Tilde{\textup{\textbf{B}}}_{\star}^{f} =  {\textup{\textbf{H}}_{\star}}(\Theta_t)^{-\frac{1}{2}} ~ \Tilde{\textup{\textbf{H}}}_{\star}^{f} ~ {\textup{\textbf{H}}_{\star}}(\Theta_t)^{-\frac{1}{2}}.
\end{equation}
We can directly establish that the matrices $\Tilde{\textup{\textbf{B}}}_{\star}^{f}$ for all $f \in \Theta_t$ satisfy
\begin{equation}\label{eqn:sum_B_mat_I_rf_trace_B_mat}
    \sum_{f \in \Theta_t}^{}\Tilde{\textup{\textbf{B}}}_{\star}^f = I_n ,\hspace{1cm}     r_f = \textup{Tr}\left( \Tilde{\textup{\textbf{B}}}_{\star}^f  \right).
\end{equation}

Let $X$ be a random matrix such that with probability
\begin{equation}\label{eqn:X_random_mat}
   p_{f} = \dfrac{\textup{Tr}({\Tilde{\textup{\textbf{B}}}_{\star}^{f}})}{\textup{Tr}(I_n)},  \hspace{1cm} X = \dfrac{\Tilde{\textup{\textbf{B}}}_{\star}^{f}}{\textup{Tr}({\Tilde{\textup{\textbf{B}}}_{\star}^{f}})}
\end{equation}



It's straightforward to show that $p_f$ has the same values as $\pi_f$ defined earlier in Remark~\ref{rem: ind_prob}). Just like $\pi_f$, $p_f$'s represent probability distribution. Computing the expected value of $X$ leads to 
\begin{equation}\label{eqn:X_expectation}
\begin{aligned}
    \mathbb{E}[X] =  \sum_{f \in \Theta_t}^{} p_f X 
    =\sum_{f \in \Theta_t}^{} \dfrac{\textup{Tr}({\Tilde{\textup{\textbf{B}}}_{\star}^{f}})}{\textup{Tr}(I_n)}\dfrac{\Tilde{\textup{\textbf{B}}}_{\star}^{f}}{\textup{Tr}({\Tilde{\textup{\textbf{B}}}_{\star}^{f}})}
    = \dfrac{I_n}{\textup{Tr}(I_n)} = \dfrac{1}{n} I_n.
\end{aligned}
\end{equation}

Let $X_1, \cdots, X_q $ be independent copies of the random matrix $X$. From equation \eqref{eqn:X_random_mat}, we can directly establish that
\begin{equation}\label{eqn:X_cone_norm}
    0 \preceq X_i \preceq I_n, \hspace{1cm} \|X_i\| \leq 1
\end{equation}
for all $i \in \{1, \dots, q\}.$
Now, we use the results from Lemma \ref{lem:matrix_conc_ineq}. 
Using \eqref{eqn:X_expectation} and \eqref{eqn:X_cone_norm}, we directly conclude that
\begin{equation}\label{eqn:R_expect_tilde_X_mu_min}
    R = 1, \hspace{0.6cm} \mathbb{E}\left[\sum_{i = 1}^{q} X_i\right] = \dfrac{q}{n} I_n, \hspace{0.6cm} \mu_{\textup{min}} = \dfrac{q}{n}.
\end{equation}
For all $0<\epsilon<1$, Lemma \ref{lem:matrix_conc_ineq} states that
\begin{equation}\label{eqn:tropp_matrix_chernoff_eq_customized}
    \scalebox{1}{$
         \mathbb{P}\left[\sum_{i = 1}^{q} X_i \succeq (1-\epsilon) \dfrac{q}{n}\right]\geq 1 - n  e^{-q\epsilon^2/2n}.
    $}
\end{equation}

Given $\kappa$ as defined in \eqref{eqn:alpha} and $\delta \in [\kappa, 3/4)$, if we choose the number of features
\[q = \dfrac{2 n}{\epsilon^2} \textup{ln}\dfrac{n}{\delta} = O(n ~ \textup{log} ~ n /\epsilon^2),\] 
it is ensured that 
\begin{equation}\label{eqn: prob_sum_Xi}
    \scalebox{1}{$
        \mathbb{P}\left[\sum_{i = 1}^{q} X_i \succeq (1-\epsilon) \dfrac{q}{n}\right] \geq 1 -\delta.
    $}
\end{equation}

We impose the restriction $\delta \geq \kappa$ in order to restrict the algorithmic complexity to $O(n ~ \textup{log} ~ n /\epsilon^2).$
From \eqref{eqn: prob_sum_Xi}, we conclude that with probability greater than $1 - \delta$, 
\begin{align*}
    \sum_{i = 1}^{q}X_i \succeq \frac{q}{n}(1-\epsilon)I_n, ~
    \sum_{f \in \Phi_t}\dfrac{\Tilde{\textup{\textbf{B}}}_{\star}^{f}}{\textup{Tr}({\Tilde{\textup{\textbf{B}}}_{\star}^{f}})} \succeq \frac{q}{n}(1-\epsilon)I_n.
\end{align*}
Define the random variable $\chi$ as 
\begin{equation}    \label{eqn:Chi_random_var}
    \scalebox{1}{$
        \zeta = \textup{inf}\left\{\gamma \bigg| ~ {\gamma} \sum_{f \in \Phi_t}\Tilde{\textup{\textbf{B}}}_{\star}^{f} \succeq \frac{q}{n}(1-\epsilon)I\right\}.
    $}
\end{equation}
From \eqref{eqn:Chi_random_var}, we directly establish that $\zeta^{-1} \geq  \underset{f \in \Phi_t}{\textup{min}} \textup{Tr}({\Tilde{\textup{\textbf{B}}}_{\star}^{f}})$. Let $\bar{\zeta} = \mathbb{E}\left[ \zeta\right]$. Then with a probability greater than $1 - \delta$,
\begin{equation}\label{eqn:chi_intro}
    \sum_{f \in \Phi_t}\Tilde{\textup{\textbf{B}}}_{t,T}^{f} \succeq \frac{q}{\zeta n}(1-\epsilon)I.
\end{equation}
Using the Markov's inequality, 
\begin{equation}\label{eqn: chi_markov}
    \mathbb{P}\left[\, \zeta \leq a \bar{\zeta} \, \right] \geq 1 - \frac{1}{a},
\end{equation}
where $a>1 - \delta.$
Let the event in \eqref{eqn:chi_intro} be denoted by $\mathfrak{A}$ and the event in \eqref{eqn: chi_markov} be denoted by $\mathfrak{B}$, then 
\begin{align*}
    \mathbb{P}\left(\mathfrak{A} \cap \mathfrak{B}\right) &= \mathbb{P}\left(\mathfrak{A}\right)  + \mathbb{P}\left(\mathfrak{B}\right) - \mathbb{P}\left(\mathfrak{A} \cup \mathfrak{B}\right)\\
    &\geq 1 - \delta + 1 - \frac{1}{a} -1\\
    &= 1 - \delta  - \frac{1}{a}.
\end{align*}
Using \eqref{eqn: prob_sum_Xi}, \eqref{eqn:chi_intro} and \eqref{eqn: chi_markov}, we conclude 
\begin{equation*}
    \scalebox{1}{$
        \mathbb{P}\left[ \sum_{f \in \Phi_t}\Tilde{\textup{\textbf{B}}}_{\star}^{f} \succeq \frac{q}{a\bar{\zeta} n}(1-\epsilon)I\right] \geq 1 - \delta  - \frac{1}{a}.
    $}
\end{equation*}
Let $a = 4$, then with probability greater than $\frac{3}{4} - \delta$, 
\[\sum_{f \in \Phi_t}\Tilde{\textup{\textbf{B}}}_{\star}^{f} \succeq \frac{(1-\epsilon)}{4\bar{\chi}}I,\]
where $\bar{\chi} = \dfrac{\bar{\zeta} n}{q}$.
Using \eqref{eqn: B_mat}, we deduce
\begin{align*}
     \sum_{f \in \Phi_t}\Tilde{\textup{\textbf{H}}}_{\star}^{f} &\succeq \frac{(1-\epsilon)}{4\bar{\chi}}~{\textup{\textbf{H}}_{\star}}(\Theta_t).
\end{align*}
Finally, using \eqref{eqn:max_info_mat} and \eqref{eqn:indiv_info_mat}, we observe
\[{\textup{\textbf{H}}}_{\star}(\Phi_t) = \sum_{f \in \Phi_t}\Tilde{\textup{\textbf{H}}}_{\star}^{f} + \dfrac{|\Theta_t| - q}{|\Theta_t|} \Tilde{\textbf{H}}_{\star},\] 
which leads us to the conclusion that
\[{\textup{\textbf{H}}}_{\star}(\Phi_t) \succeq \frac{(1-\epsilon)}{4\bar{\chi}}{\textup{\textbf{H}}_{\star}}(\Theta_t),\] 
holds with probability atleast $3/4 - \delta.$\\

\vspace{1mm}
\noindent \underline{\bf Proof of Theorem \ref{thm:leverage_score_monotonicity}:}
For simplicity of notation, let us define 
\[X = \Tilde{\textbf{H}}_{\star}, \hspace{0.3cm} A_0 =  \sum_{f\in \Theta_t}^{} \textbf{H}_{\star}^f, \hspace{0.3cm} B_0 = {{\textbf{H}_{\star}^f}}, \hspace{0.3cm} a_0 =  \dfrac{1}{|\Theta_t|}.\]
Using definition \ref{def:lever_score}, 
\begin{equation}\label{eqn:leverage_score_alt}
    r_f = \mathrm{Tr}\left((X+A_0)^{-1}(a_0 X+ B_0)\right),
\end{equation}

where it is straightforward to verify that $A_0, B_0 \succeq 0$ and $a_0 >0$. If $B_0 \succeq a_0 A_0$, then $r_f$ is monotonically decreasing in $X$. If  $B_0 \preceq a_0 A_0$, then $r_f$ is monotonically increasing in $X$. 
We can rewrite \eqref{eqn:leverage_score_alt} as function of $X$ such that

\begin{equation}\label{eqn:leverage_score_alt_2}
    r_f(X) = \mathrm{Tr}\left((X+A_0)^{-1}(a_0 X+ B_0)\right),
\end{equation}



For the first case, let us assume that $B_0 \succeq a_0 A_0$. 
Using \eqref{eqn:team_relative_meas_info_mat_t_M}, \eqref{eqn:inform_matrix_update_rel_meas},and \eqref{eqn:info_matrix_update_feature_f}, one can show that
\[\mathcal{L}_{t:t+M} \preceq \mathcal{L}'_{t:t+M} \implies X \preceq X'\]

Then, we have $(X' + A_0)^{-1} \preceq (X + A_0)^{-1}$, since the inverse preserves the order for positive definite matrices. Since $B_0 \succeq \alpha A$, it implies $B_0 - a_0 A_0 \succeq 0$. The trace of a product of positive semi-definite matrices is non-negative, thus we have 
\[ \mathrm{Tr}((X' + A_0)^{-1}(B_0 - a_0 A_0)) \leq \mathrm{Tr}((X + A_0)^{-1}(B_0 - a_0 A_0)). \]
This implies  $r_f' \leq r_f$, where $r'_{f}$ and $r_{f}$ are leverage scores when network information matrices are $\mathcal{L}'_{t:t+M}$ and $\mathcal{L}_{t:t+M}$, respectively. Therefore, $r_f$ is monotonically decreasing when $B_0 \succeq a_0 A_0$.


For the second case, let us assume that $B_0 \preceq a_0 A_0$. Let $X, X' \in \mathcal{S}^n_+$ such that $X \preceq X'$. As in the previous case, we have $(X' + A_0)^{-1} \preceq (X + A_0)^{-1}$. Since $B_0 \preceq a_0 A_0$, it implies $B_0 - a_0 A_0 \preceq 0$. However, the trace operator is linear and preserves order for positive semi-definite matrices. Thus, we have 
\[ \mathrm{Tr}((X' + A_0)^{-1}(B_0 - a_0 A_0)) \geq \mathrm{Tr}((X + A_0)^{-1}(B_0 - a_0 A_0)). \]
This implies $r'_f \geq r_f$. Thus,  $r_f$ is monotonically increasing when $B_0 \preceq a_0 A_0$.

\end{document}